# Improving Cone-Beam CT Image Quality with Knowledge Distillation-Enhanced Diffusion Model in Imbalanced Data Settings

Joonil Hwang[1,2]*, Sangjoon Park[4]*, NaHyeon Park[3],

Seungryong Cho[1,2+], Jin Sung Kim[4,5,6+]

[1] Department of Nuclear and Quantum Engineering, Korea Advanced Institute of Science and Technology, Daejeon, Republic of Korea
[2] Medical Image and Radiotherapy Lab` (MIRLAB), Korea Advanced Institute of Science and Technology, Daejeon, Republic of Korea
[3] Department of Artificial Intelligence, Korea Advanced Institute of Science and Technology, Seoul, Republic of Korea
[4] Department of Radiation Oncology, Yonsei Cancer Center, Yonsei University College of Medicine, Seoul, Republic of Korea
[5] Medical Physics and Biomedical Engineering Lab (MPBEL), Yonsei University College of Medicine, Seoul, Republic of Korea
[6] OncoSoft, Seoul, Republic of Korea

*These authors contributed equally to this study.
[+]These authors contributed equally and are co-corresponding authors to this study.

**Abstract**. In radiation therapy (RT), the reliance on pre-treatment computed tomography (CT) images encounter challenges due to anatomical changes, necessitating adaptive planning. Daily cone-beam CT (CBCT) imaging, pivotal for therapy adjustment, falls short in tissue density accuracy. To address this, our innovative approach integrates diffusion models for CT image generation, offering precise control over data synthesis. Leveraging a self-training method with knowledge distillation, we maximize CBCT data during therapy, complemented by sparse paired fan-beam CTs. This strategy, incorporated into state-of-the-art diffusion-based models, surpasses conventional methods like Pix2pix and CycleGAN. A meticulously curated dataset of 2800 paired CBCT and CT scans, supplemented by 4200 CBCT scans, undergoes preprocessing and teacher model training, including the Brownian Bridge Diffusion Model (BBDM). Pseudo-label CT images are generated, resulting in a dataset combining 5600 CT images with corresponding CBCT images. Thorough evaluation using MSE, SSIM, PSNR and LPIPS demonstrates superior performance against Pix2pix and CycleGAN. Our approach shows promise in generating high-quality CT images from CBCT scans in RT.

## 1 Introduction

Sole reliance on pre-treatment CT scans risks under-dosing tumors and exposing organs at risk (OARs) to excessive radiation due to anatomical changes during treatment. Adaptive radiation therapy (ART) enhances patient outcomes by safeguarding healthy tissues and optimizing tumor doses, leveraging daily CBCT for real-time adaptability. Yet, CBCT's basic accuracy in tissue density and contrast falls short of CT standards [1]. Typically, planning CT (pCT) is adapted to CBCT anatomy using deformable image registration (DIR) for accurate dose calculations, maintaining pCT's Hounsfield Unit (HU) precision. However, the anatomical fidelity of deformed pCT and potential inaccuracies from DIR, exacerbated by significant anatomical shifts and reduced contrast in soft tissue, pose challenges [2].

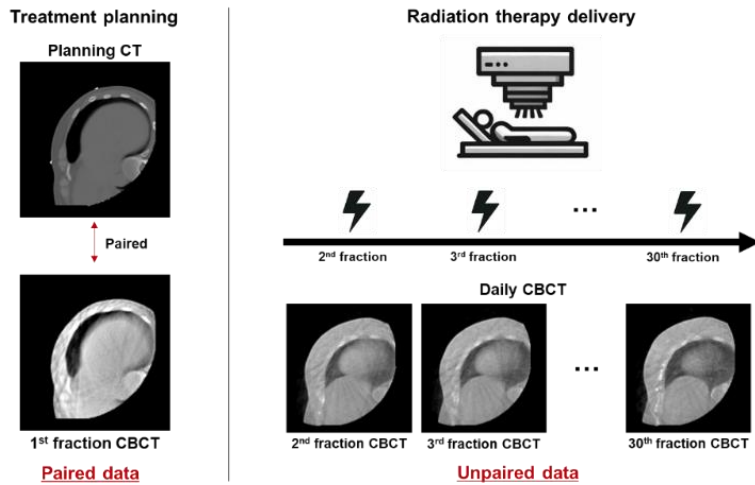

**Figure 1.** Illustration of real-world image acquisition settings for radiation therapy

Efforts have been ongoing to generate pseudo-CT images from CBCT that are comparable in quality to pCT images without the need for additional pCT acquisitions [14,15]. In typical radiation therapy delivery, as depicted in Figure 1, pCT is acquired only 2-3 times for treatment planning and considering ART, whereas CBCT is captured daily. This results in a data imbalance, with only 2-3 pairs of pCT-CBCT data per patient against a larger volume of unpaired CBCT images. A primary option to address this issue is supervised learning, which trains models on paired CBCT and pseudo-CT data, but this method is limited to using only a fraction of the total data [11]. Alternatively, some studies have proposed using unsupervised domain-to-domain translation techniques to synthesize pCT-like images from the abundant unpaired CBCT and pCT data [16]. However, these unsupervised learning approaches often yield suboptimal performance in terms of tissue density accuracy due to the inherent limitations of unsupervised learning methods.

To address this issue, we propose a novel method utilizing a self-training framework with knowledge distillation for the CBCT to pCT conversion task. This approach maximizes the utilization of a small number of paired CBCT-pCT data and a large amount of CBCT data, surpassing both traditional supervised methods relying solely on paired data and unsupervised baselines utilizing unpaired data. We present and validate this new method through experiments.

The proposed method in this study facilitates real-time adjustments to radiation therapy plans by providing high-quality CBCT images that closely match the patient's current anatomical structure, accommodating changes in tumor size and position.

## 2 Methodology

### 2.1 Background of Brownian Bridge Diffusion Model (BBDM)

The Brownian Bridge diffusion process [5] follows a distinct trajectory compared to the De-noising Diffusion Probabilistic Model [6] (DDPM). Instead of converging to a purely Gaussian noise distribution, it concludes by reaching the clean conditional input denoted as 'Q.' Employing similar notations as DDPM, where (P, Q) represents paired training data from the CBCT and CT domains. The forward diffusion process of the Brownian Bridge is defined as follows:

$$q_{Brownian\ Bridge}(P_t|P_0, Q) = N(P_t; (1 - k_t)P_0 + k_t Q, \sigma_t I) \quad (1)$$

Here, $P_0 = P, k_t = \frac{t}{T}$ (with T being the total steps of the diffusion process), and $\sigma_t$ is the variance. The transition probability $q_{Brownian\ Bridge}(P_t|P_{t-1}, Q)$ is derived as follows:

$$q_{Brownian\ Bridge}(P_t|P_{t-1}, Q) = N\left(P_t; \frac{1-k_t}{1-k_{t-1}}P_{t-1} + \left(k_t - \frac{1-k_t}{1-k_{t-1}}k_{t-1}\right)Q, \sigma_t - \sigma_{t-1}\frac{(1-k_t)^2}{(1-k_{t-1})^2}I\right) = \left((1-k_t)P_0 + k_t Q + \sqrt{2(k_t - k_t^2)}\delta_t; \frac{1-k_t}{1-k_{t-1}}((1-k_{t-1})P_0 + k_t Q + \sqrt{2(k_{t-1} - k_{t-1}^2)}\delta_{t-1}) + \left(k_t - \frac{1-k_t}{1-k_{t-1}}k_{t-1}\right)Q, (2(k_t - k_t^2) - 2(k_{t-1} - k_{t-1}^2))\frac{(1-k_t)^2}{(1-k_{t-1})^2}I\right) \quad (2)$$

Here, $\delta_t, \delta_{t-1} \sim N(0, I)$. As per Eq. (2), when the diffusion process reaches the destination (t = T), $k_t = 1$, and $P_T = Q$. The forward diffusion process establishes a consistent mapping from domain CBCT to CT.

In contrast to the reverse processes of conventional diffusion models, the typical procedure initiates with extracting pure noise from a Gaussian distribution and gradually eliminates noise to achieve a clean data distribution. Existing approaches involve incorporating the condition as an additional neural network input during the reverse diffusion process to model the conditional distribution.

Conversely, the proposed Brownian Bridge process adopts a unique approach by directly commencing from the conditional input, where $P_T = Q$. Aligned with the fundamental concept of denoising diffusion methods, our method's reverse process aims to predict $P_{t-1}$ based on $P_t$:

$$p_\theta(P_{t-1}|P_t, Q) = N(P_{t-1}; \mu_\theta(P_t, t), \bar{\sigma}_t I) \qquad (3)$$

Here, $\mu_\theta(x_t, t)$ represents the predicted mean value of noise, and, $\bar{\sigma}_t$ denotes the noise variance at each step. Similar to DDPM, the mean value $\mu_\theta(P_t, t)$ is learned through a neural network with parameters $\theta$ based on the maximum likelihood criterion. While the variance, $\bar{\sigma}_t$ does not undergo a learning process, its analytical form significantly influences high-quality image translation.

Importantly, the reference image Q from the CT domain exclusively serves as the starting point $P_T = Q$ for the reverse diffusion. Notably, it is not utilized as a conditional input in the prediction network $\mu_\theta(P_t, t)$ at each step.

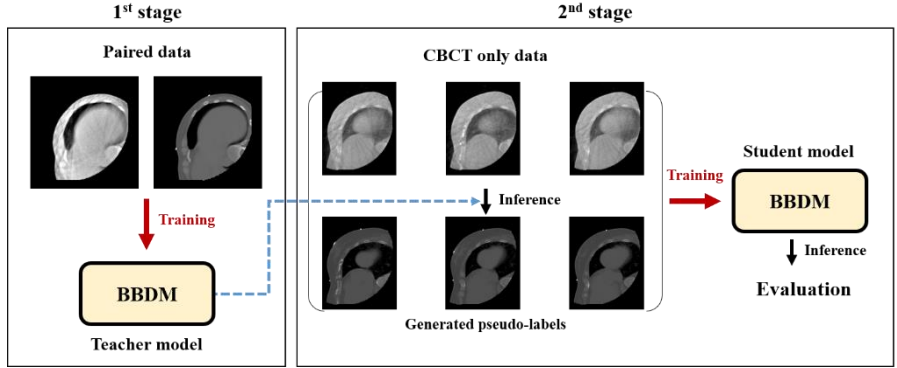

**Figure 2. Overall workflow of the proposed method**

Through this approach, it can exhibit superior performance in image-to-image translation compared to not only GAN-based approaches [3,4] with training instability and mode collapse issues but also other diffusion models designed for different conditional generation tasks.

### 2.2 Knowledge distillation-based self-training

Knowledge distillation [7] is a model compression technique that involves transferring insights from a complex teacher model to a simpler student model. The primary objective is to distill valuable knowledge and generalization capabilities embedded in the teacher model, delivering them to a more compact and computationally efficient student model. Typically, a large and high-performing neural network serves as the teacher model, acting as the initial source of information. The student model, often smaller in size and computational complexity, is trained to replicate not only the output predictions of the teacher model but also its internal representations and ac-quired knowledge.

Proven to be particularly beneficial in resource-constrained scenarios, such as edge devices or mobile applications, knowledge distillation contributes to enhanced model generalization and robustness.

Knowledge distillation is also employed to enhance model performance. In the context of Self-training with a noisy student [8], knowledge distillation is utilized in scenarios where limited labeled data is available, while unlabeled data is abundant, aiming to improve model performance. In this approach, the teacher model is initially trained with labeled data, and subsequently, the student model is trained, incorporating the generated pseudo-labels from the teacher model. This recursive process, wherein the student model becomes a new teacher progressively, demonstrates the incremental enhancement of model performance. Notably, introducing noise in various ways during the learning process enhances the resulting model's robustness. Motivated by this paradigm, we apply a similar framework in a real clinical setting where CBCT-pCT pairs are limited, yet CBCT data is abundant, with the objective of improving performance.

### 2.3 Preprocessing

For preprocessing, alignment and cropping procedures were employed to ensure coherence between the CBCT and CT volumes, with the CT slices adjusted to match the region of interest in the corresponding CBCT scans. Subsequently, the entire dataset underwent normalization to the standardized range of [-1, 1].

### 2.4 Proposed Framework

In the initial phase, the BBDM served as the foundational teacher model, trained on 1400 paired CBCT and CT scans. This teacher model was then employed to generate pseudo-labeled synthesized pCT images from 4200 unpaired CBCT images. In the subsequent phase, a secondary student model was trained, initialized with the weights from the teacher model, leveraging both the 1400 paired and the 4200 pseudo-labeled datasets. The performance of this secondary student model was benchmarked against various baseline models and the original teacher model to assess the effectiveness of our approach. Figure 2 graphically represents the entire training framework.

### 2.5 Evaluation metrics

We conducted a thorough evaluation encompassing normalized CT and CBCT images alongside the outputs generated by our proposed models. To quantitatively analyze the results, we employed three metrics: Mean Square Error (MSE), Structural Similarity Index Measure (SSIM), Peak Signal-to-Noise Ratio (PSNR) and Learned Perceptual Image Patch Similarity (LPIPS). These metrics were selected to assess both the accuracy and structural fidelity of the generated CT images in comparison to the ground truth pCT images.

## 3 Experiments

### 3.1 Datasets

In this study, we analyzed data from a cohort of 100 breast cancer patients who underwent adjuvant RT following breast-conserving surgery between March 2020 and December 2021. We curated a dataset comprising 1400 pairs of CBCT and CT scans from 40 breast cancer patients, supplemented by an unpaired 4200 CBCT scans from 60 patients, emulating the real clinical situation for RT. Planning CT images were obtained before the initiation of treatment for all patients, utilizing a TOSHIBA scanner operating at 120 kVp, with a voxel size of 1.367 mm × 1.367 mm × 3 mm, and a voxel count of 512 × 512 × (range, 122–179).

For each treatment session, daily CBCT images were acquired using an ELEKTA XVI scanner, employing a voltage of 100 kVp, voxel size of 1.367 mm × 1.367 mm × 3 mm, and a voxel count of 512 × 512 × (range, 122–179). The CBCT scans underwent automatic registration to the pCT system, employing an algorithm accounting for three degrees of freedom and the grey values of the images.

Following the initial registration process, manual adjustments were performed as needed to ensure precise alignment between the CBCT and pCT scans. This registration process was crucial to guarantee accurate spatial alignment, facilitating subsequent analyses and comparisons between the two imaging modalities.

## 4 Model Comparison

We compared our model with the supervised model (Pix2pix) [9] as well as the unsupervised model (CycleGAN) [10]. For Pix2pix, we utilized a paired dataset consisting of 1400 slices each of CBCT and CT. For CycleGAN, the full dataset was utilized in an unsupervised approach, encompassing 5600 CBCT and 1400 CT scans. The evaluation then proceeded with an independent set of 1400 paired CBCT and labeled CT images from a distinct patient cohort.

**Table. 1.** Quantitative assessment of the input, pix2pix, CycleGAN, Teacher model, and ours in terms of MSE, SSIM, PSNR, and LPIPS. **Bold** notes the best results

| Metric | Input | Pix2pix | CycleGAN | Teacher model | Student model (proposed) |
|---|---|---|---|---|---|
| MSE | 0.0513 | **0.0047** | 0.0179 | 0.0048 | **0.0047** |
| SSIM | 0.7783 | 0.8946 | 0.8742 | 0.9180 | **0.9210** |
| PSNR | 13.03 | 23.69 | 18.08 | 23.74 | **23.79** |
| LPIPS | 1.762e-4 | **2.074e-5** | 3.074e-5 | 2.144e-5 | 2.158e-5 |

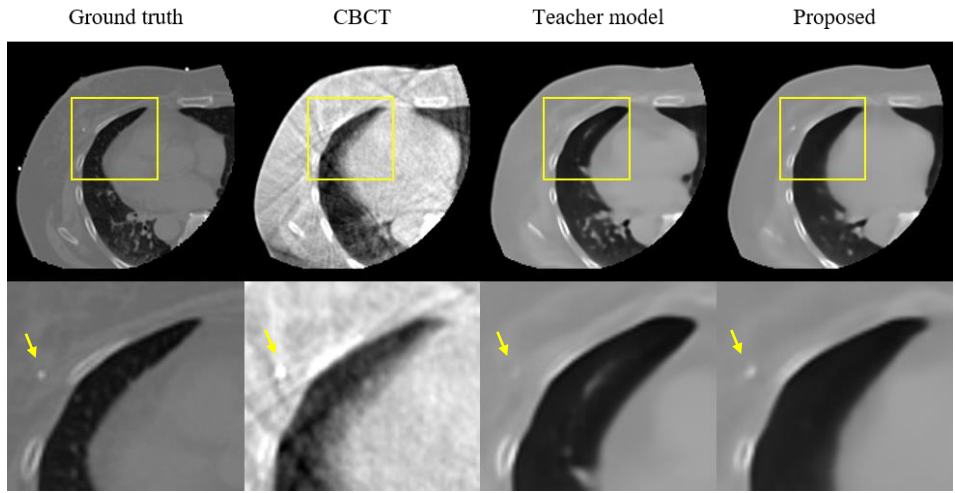

**Figure 3.** Comparison between ground truth (pCT), CBCT, teacher model and the proposed method.

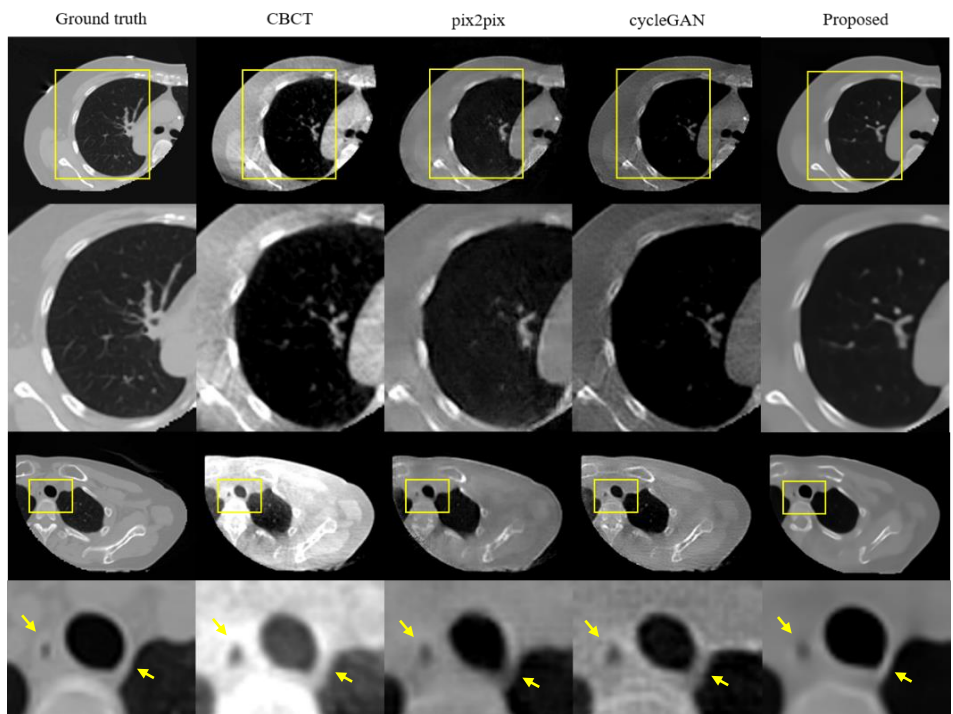

**Figure 4.** Comparison between ground truth (pCT), CBCT, pix2pix, cycleGAN, and the proposed method.

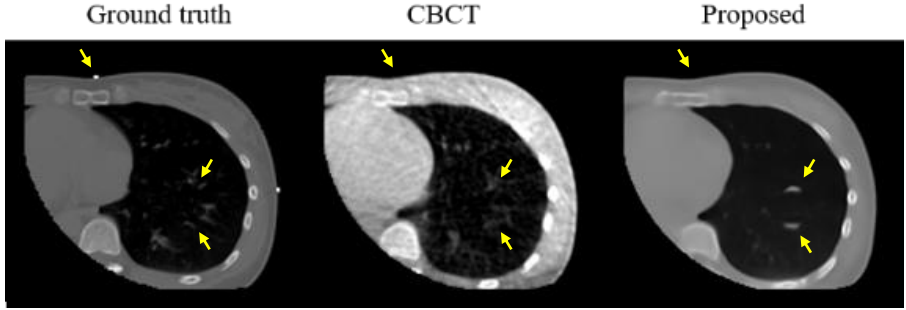

**Figure 5.** Exemplified failure cases of the proposed method

## 5 Results

### 5.1 Quantitative Results

Quantitatively, the proposed method outperformed all counterparts in reducing streak artifacts, as substantiated by superior SSIM and PSNR values presented in Table 1. While Pix2pix showed competitive performance with a lower MSE and slightly better results in the LPIPS metric, our approach consistently demonstrated closer alignment to the HU of the ground truth, underlining its efficacy in artifact reduction. Comparative experiments against baseline models show exceptional outcomes in MSE, PSNR, and SSIM, even surpassing scenarios using only paired data. Despite a slight decrease in LPIPS, overall metrics indicates the superior performance of the student model obtained with knowledge distillation.

### 5.2 Qualitative Results

As shown in the Figure 3 and 4, the proposed method notably reduced streak artifacts while preserving the image fidelity closer to the ground truth (pCT) images. The qualitative comparison shows the proposed method's superior ability to maintain the clarity and detail of the original images. This is in contrast to Pix2pix, which, despite its reasonable quantitative performance, tended to produce overly blurry images, leading to visual distortions.

### 5.3 Failure Cases

Figure 5 shows a failure case. While the method successfully adjusted the overall HU closer to those of pCT, it also introduced non-existent structures, as depicted. This phenomenon is indicative of generative artifacts, a common challenge in generative models. The issue is especially prominent in small structures, such as the small bronchi.

Conversely, when structures that were not clearly visible in CBCT, the elimination of some structures were noted.

## 6 Conclusion

In conclusion, CBCT to pCT conversion faces challenges such as dependencies on re-planning CT, discrepancies in HU accuracy within CBCT datasets, and differences in organ shapes and locations between CT and CBCT [11-13]. Our method, combining supervised and unsupervised approaches, shows superior performance in MSE, PSNR, and SSIM by utilizing both paired and unpaired data. Although our study makes significant advancements over baseline models, certain limitations remain.

This study underscores the potential for refining ART plans through accurate pCT image generation. Future research will focus on ensuring generalizability across different CT scanners and diverse patient populations, exploring advanced data normalization techniques and multi-frequency processing.

Our primary aim was to enhance daily CBCT image quality for better patient positioning and treatment monitoring. For these purposes, precise HU accuracy is less critical. The proposed model effectively aids in treatment monitoring and adaptive planning, providing clinical utility without direct dose calculations.

**Acknowledgments.** None.

**Disclosure of Interests.** The authors have no competing interests to declare that are relevant to the content of this article.